\documentclass[10pt,twocolumn,letterpaper]{article}

\usepackage[pagenumbers]{cvpr} 

\usepackage{graphicx}
\usepackage{amsmath}
\usepackage{amssymb}
\usepackage{booktabs}
\usepackage{enumitem}
\graphicspath{ {./Figures/} }

\usepackage[pagebackref,breaklinks,colorlinks]{hyperref}

\usepackage[capitalize]{cleveref}
\crefname{section}{Sec.}{Secs.}
\Crefname{section}{Section}{Sections}
\Crefname{table}{Table}{Tables}
\crefname{table}{Tab.}{Tabs.}

\def\cvprPaperID{9} 
\def\confName{CVPR}
\def\confYear{2022}

\begin{document}

\title{Concept Activation Vectors for Generating User-Defined 3D Shapes}

\author{Stefan Druc$^1$, Aditya Balu$^2$, Peter Wooldridge$^1$, Adarsh Krishnamurthy$^2$, Soumik Sarkar$^2$\\
{\small $^1$MonolithAI - 12-18 Hoxton Street, London N1 6NG, UK}\\
{\small $^2$Iowa State University - Ames, Iowa, USA 50011}\\
{\tt\small $^1$\{stefan,peter\}@monolithai.com, $^2$\{baditya,adarsh,soumiks\}@iastate.edu}
}

\maketitle

\begin{abstract}
We explore the interpretability of 3D geometric deep learning models in the context of Computer-Aided Design (CAD). The field of parametric CAD can be limited by the difficulty of expressing high-level design concepts in terms of a few numeric parameters. In this paper, we use a deep learning architectures to encode high dimensional 3D shapes into a vectorized latent representation that can be used to describe arbitrary concepts. Specifically, we train a simple auto-encoder to parameterize a dataset of complex shapes. To understand the latent encoded space, we use the idea of Concept Activation Vectors (CAV) to reinterpret the latent space in terms of user-defined concepts. This allows modification of a reference design to exhibit more or fewer characteristics of a chosen concept or group of concepts. We also test the statistical significance of the identified concepts and determine the sensitivity of a physical quantity of interest across the dataset.
\end{abstract}

\section{Introduction}
\label{sec:intro}
\begin{figure*}[th!]
  \centering
  \includegraphics[width=0.9\linewidth]{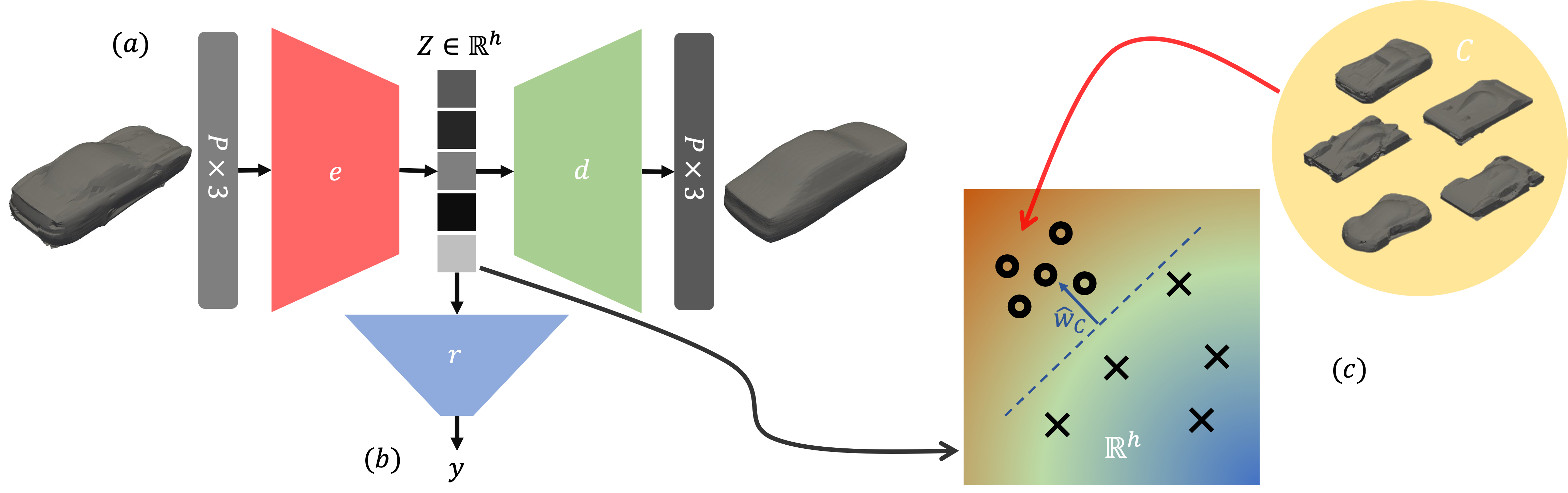}
  \caption{Pictorial representation of our method. (a) We train an auto-encoder to reconstruct point clouds of car shapes, (b) we then train another network to predict the drag coefficient from the latent space, $Z$. (c) Finally we collect examples from the dataset to define human interpretable concepts and then represent them in the latent space of the auto-encoder .}
  \label{fig:method}
\end{figure*}
Over the past half-decade, there has been spectacular success with deep learning to synthesize 3D shapes~\cite{Yin2018P2PNET,Park2019DeepSDFLC,Achlioptas2017RepresentationLA,Chen2019LearningIF,Mildenhall2020NeRFRS,Mller2022InstantNG,chen2020bsp,deng2020cvxnet,paschalidou2021neural}. A common approach has been to use a latent space representation that can encode the 3D geometry using some non-linear transformations. This latent space can either be learned directly from the data through an auto-encoder architecture \cite{Yin2018P2PNET,Guan2020GeneralizedAF} or implicitly by sampling from a known distribution (e.g., Gaussian)~\cite{Donahue2017AdversarialFL,Wu2016LearningAP,Chen2019LearningIF,Achlioptas2017RepresentationLA,Nash2017TheSV} and training a decoder network to replicate the samples from the dataset. Other latent representation approaches also include multi-view reconstruction~\cite{Mildenhall2020NeRFRS,Mller2022InstantNG,wu20153d,su15mvcnn,song2016deep,li2018sonet,kanezaki2016rotationnet,SFIKAS2018208,qi2016volumetric}, implicit neural representations~\cite{Sitzmann2020ImplicitNR,atzmon2020sal,atzmon2020sald,gropp2020implicit}, auto-decoder models~\cite{Park2019DeepSDFLC}, etc. The general aim in all these approaches is to obtain a latent space whose dimension is much smaller than the input dimension, thus having a strong relation to the field of signal compression. The latent space representation has been interpreted as a deep learning analog of dimensionality reduction algorithms such as SVD and PCA; indeed, a network with a single hidden layer can be made equivalent to the $\mathbf{U\Sigma V}$ formulation of SVD~\cite{lecun2015deep}. Thus, it is believed that the learned latent space should encode the most salient modes of the data and forms a low-dimensional manifold. Similar to their classical analogs, deep learning models discover the latent manifold in a self-supervised manner, where the objective function is determined by recreating the input distribution. Furthermore, either by construction or assumption, it is possible to linearly interpolate between points on this manifold and obtain synthesized data points. To this end, some works impose regularising conditions on the latent space to encourage the network to discover a well-behaved manifold, such as sparsity constraints \cite{Lin20163DKD} or reparametrization to multi-variate Gaussian distributions \cite{Brock2016GenerativeAD}.

One fundamental drawback of the latent space representation is its lack of interpretability. In the standard approach, there is no control over what each coordinate in the latent vector represents, and there is no canonical way in which the network can decompose the original signal. It is easy enough to encode two dataset samples and linearly interpolate between them; however, it is much harder to determine the changes required apriori. There have been attempts to condition the embedding space on a particular semantic feature in the input \cite{Kudo2019VirtualTS}, thus allowing more principled exploration. However, these conditions need to be defined before training, thus restricting the range of possible semantic combinations that can be investigated. In the engineering context it would be highly desirable to link high level design concepts with physical performance. Usually this is done through simulations for an individual design, but quantifying the impact of a design choice across an entire database is a challenging task. We show that using the representation power of deep neural networks we can analyse the sensitivity of a physical quantity of interest with respect to user-defined design concepts and across an entire dataset.

The idea of explainable AI (XAI) is also emerging over the past half-decade. Very early approaches for XAI were gradient-based saliency maps~\cite{simonyan2013deep}, class-activation maps~\cite{zhou2016learning,selvaraju2017grad}, and some recent approaches are Layerwise relevance propagation~\cite{binder2016layer}. In the field of geometric shape understanding, several researchers have attempted to extend these interpretability algorithms to 3D geometric shapes represented by voxels~\cite{ghadai2018learning,yoo2021explainable} or,  more recently, point clouds~\cite{tan2022surrogate,su2021learning,zhang2019explaining,huang2019claim}. However, these methods were mostly characterized by visual appearance and, in many situations, did not satisfying some of the sanity checks outlined in ~\cite{adebayo2018sanity,tomsett2020sanity,kim2018interpretability,adebayo2020debugging}. Further, to the best of the knowledge of the authors, there are no XAI-based approaches for understanding the latent shape representation of the geometries. 

An XAI method robust to these sanity checks with a lot of promise is Concept Activation Vectors (CAVs)~\cite{kim2018interpretability,goyal2019explaining}. CAVs allow the definition of semantic representations on the embedding space in a posthoc manner. It offers the freedom to define and explore the latent manifold in a human interpretable way, with concepts and inputs that were not originally part of the training dataset. In this work, we focus on whether we can use CAVs for 3D datasets. To this end, we eschew some of the more involved representation learning techniques and propose to learn a simple MLP auto-encoder on a point cloud dataset. We chose the dataset to contain shapes that are low resolution enough to be efficiently learned by an MLP and varied enough to allow interesting interpolations. Once the learning has converged for the reconstruction loss, we explore the latent space in terms of several hand-defined concepts. To create the CAVs, we gather examples within the dataset and from a synthetic distribution that can be interpreted as abstract examples of a style. We then show that it is possible to take a reference design and add or subtract varying amounts of each concept, allowing a potential designer to modify an existing design in a high-level, human interpretable way. To further take advantage of the XAI properties of TCAVs and test the statistical significance of the learned concepts, we also train a model to predict a physical quantity from the latent space. 
\newpage
The main contributions of our work are:
\begin{itemize}[nosep,topsep=3pt]
    \item We introduce concept activation vectors (CAV) for \\interpretable 3D shape interpolation.
    \item We demonstrate CAVs for a dataset of 3D cars similar to ShapeNet.
    \item We introduce the notion of parametric concepts to parametrize designs using human-understandable concepts and regress through them.
\end{itemize}
While these contributions apply to general 3D shape understanding, we are mainly interested in an engineering design perspective. Hence, we chose a cars dataset and its corresponding drag coefficients in our experiments.

\section{Concept Exploration for 3D shapes}
\label{sec:3d_concept}

In this section, we outline the various steps required to traverse a latent space in a human interpretable manner.

\subsection{Training an auto-encoder}

We start with a dataset of point clouds $\mathcal{D}_x = \{x_i\}_{i=1}^N$ where each point cloud $x_i \in \mathbb{R}^{P\times d}$ is a rank two tensor signifying a collection of points in 3D space with $d$ features ($d=3$). In addition we can consider a set of vector quantities $\mathcal{D}_y = \{y_i\}_{i=1}^N$ related to the the point clouds, where $y_i \in \mathbb{R}^F$. We signify the combination of the two datasets as $\mathcal{D}_{x,y} = \{(x_i, y_i)\}_{i=1}^N$. An auto-encoder can be represented as a pair of maps $AE := (e,d)$ such that: 
\begin{equation}
  e : x \in \mathbb{R}^{P\times d} \to z \in Z
  \label{eq:encoder}
\end{equation}
and 
\begin{equation}
  d : z \in Z \to x \in \mathbb{R}^{P\times d}
  \label{eq:decoder}
\end{equation}
The maps $e$ and $d$ are known as the encoder and decoder, respectively. $Z \subset \mathbb{R}^h$ is referred to as the $h$-dimensional latent space, and it represents the salient manifold of the data. In practice we construct the auto-encoder as a pair of deep neural networks $AE_{\Theta}:= (e_{\theta_e},d_{\theta_d})$ where $\Theta = \{\theta_e,\theta_d\}$ is the set of parameters from all networks. We train in an end-to-end fashion with the reconstruction loss:
\begin{equation}
  \mathcal{L}(AE_{\Theta},x_i) := \left \|x_i - d_{\theta_d}(e_{\theta_e}(x_i)) \right \|^2_F;
  \label{eq:loss}
\end{equation}
where $\left \|\cdot \right \|_F$ denotes the Frobenius norm. We want to minimize the expectation of the loss across the dataset, for which we use stochastic gradient descent based optimization.  

\subsection{TCAV framework}

Once the $AE$ is trained, we can use the encoder to obtain latent vectors for any input that matches the dimensions of our original dataset. These new inputs are usually examples from a holdout dataset that follows the same distribution as $\mathcal{D}_x$. However, as was shown in \cite{kim2018interpretability}, we are not entirely limited by the input distribution, and we can obtain useful embeddings for shapes that are similar to the original input as well. We define a concept as a human interpretable collection of shapes that have some quality in common (e.g., a set of sporty cars).
\begin{equation}
    C := \{z^{(c)}: z^{(c)} = e(x^{(c)}), x^{(c)} \in \mathbb{R}^{P\times d}\}
    \label{eq:concept}
\end{equation}

Conversely, we define a non-concept $\overline{C}$ as a random collection of inputs that have no particular common characteristic. We can use concepts and non-concepts to define multi-way linear classification problems. Since linear classification learns hyperplanes that maximally separate the classes, we can take the normal to the hyperplanes to represent the direction of the concept.
\begin{equation}
    w_C = \max_{w \in \mathbb{R}^h} \Big [ \underset{z \sim C}{\mathbb{E}}\big(\mathbb{I}_{w \cdot z +b \geq 1}\big) + \underset{z \nsim C}{\mathbb{E}}\big(\mathbb{I}_{w \cdot z +b \leq -1}\big)\Big ]
    \label{eq:CAV}
\end{equation}
Here $\mathbb{I}_A$ is the indicator function, and $w_C$ are known as Concept Activation Vectors (CAVs). We would like to note that since CAVs are obtained as a result of a classification problem, they are sensitive to the other classes. For example, obtaining the CAV for concept $C_1$ against a non-concept $\overline{C}$ is not guaranteed to be the same as the CAV obtained by classifying against another concept $C_2$. In fact, this is a useful property that allows us to obtain relative and more targeted CAVs. 

\noindent\textbf{Testing with CAVs}. Once we have recovered the CAV for our chosen concept, it is easy to determine the sensitivity of a network output to the chosen concept by performing the directional derivative:
\begin{equation}
    S^C_{ij}(x) = w_C \cdot \mathbf{\nabla}_{z} d(e(x))_{ij}
    \label{eq:sensitivity}
\end{equation}
Having defined the sensitivity of a single input, we can extend it to test the sensitivity of an entire dataset or a subset of it.
\begin{equation}
    TCAV^C_{ij}(\mathcal{D}_x) := \frac{\left| \{x \in \mathcal{D}_x : S^C_{ij}(x) >0 \} \right |}{\left| \mathcal{D}_x \right|}
    \label{eq:TCAV}
\end{equation}
This gives us the fraction of tested inputs positively influenced by the concept. As mentioned in \cite{kim2018interpretability}, it is possible to consider a metric based on the magnitude of the sensitivities as well.

\subsection{Regressing from the latent space}

Since understanding the sensitivity of the auto-encoder output is not particularly interesting, we train another neural network $r_{\theta_r}$ that can regress a quantity of interest from the latent space.
\begin{equation}
    r:Z \to y
\end{equation}
Since this is a scalar physical quantity $y \in \mathbb{R}$, the sensitivities have a more intuitive interpretation than for the auto-encoder case:
\begin{equation}
    S^C_{y}(x) = w_C \cdot \mathbf{\nabla}_{z} y(z).
    \label{eq:sensitivity_y}
\end{equation}
That is, the sensitivity captures how much the quantity $y$ increases or decreases if we move in the direction of the concept. This value, along with the corresponding TCAV metric, can have practical engineering applications. 

\subsection{Exploring the latent space with CAVs}
\label{sec:exploring}

\begin{figure}[b!]
  \centering
  \includegraphics[width=0.65\linewidth]{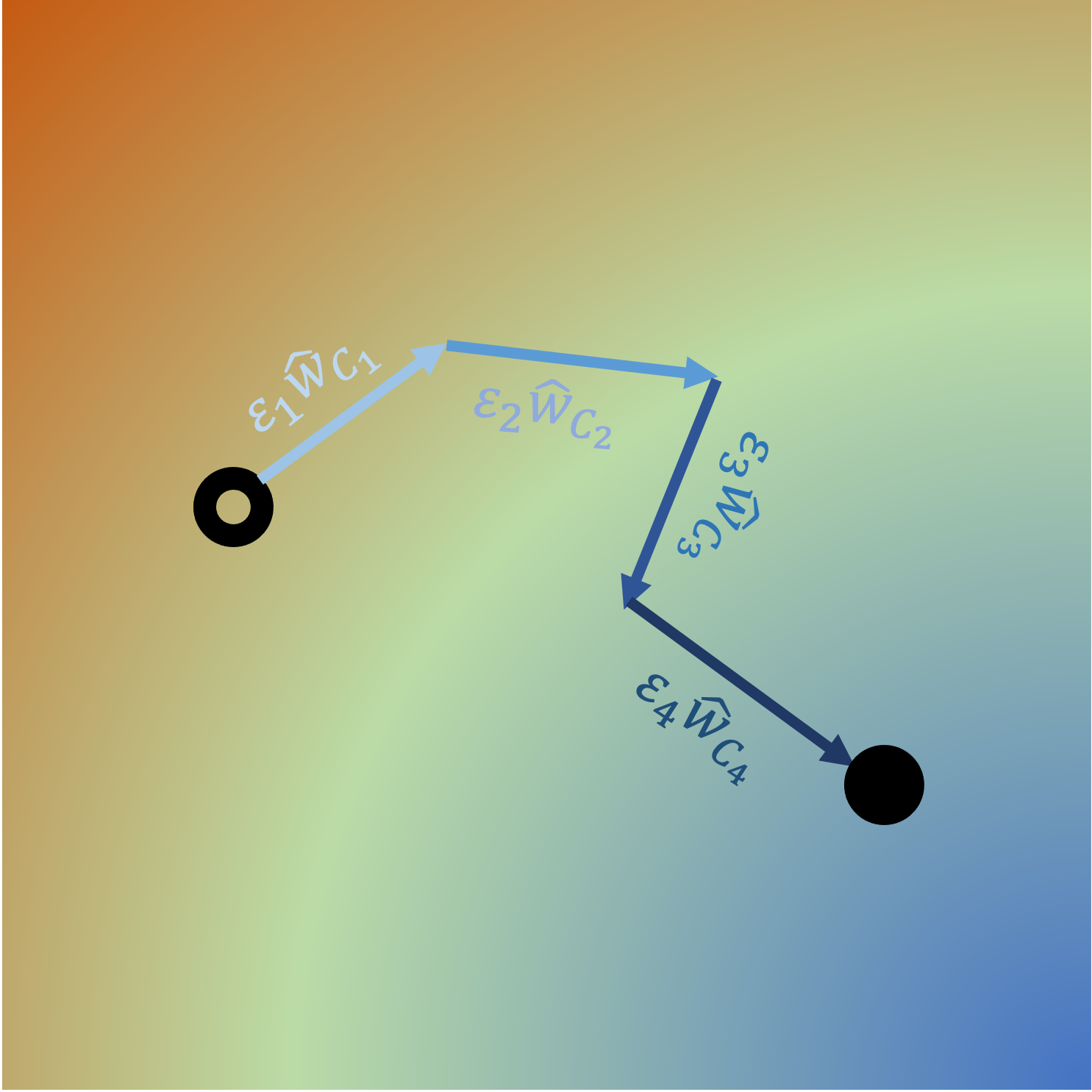}
  \caption{A conceptual path in latent space.}
  \label{fig:latent_trans}
\end{figure}

Since each CAV describes an un-normalized direction in latent space, we can use them to translate a point in $z \in \mathbb{R}^h$ along this direction:
\begin{equation}
    z' = z + \varepsilon\hat{w}_C,
\end{equation}
where $\varepsilon \in \mathbb{R}$ is a parameter controlling how far we go towards the concept if $\varepsilon > 0$ or away from the concept if $\varepsilon < 0$.  Then we can use the decoder to transform the new point in latent space back into the input space $x' = d(z')$. In this way, we can both navigate the latent space and interpret the output in a naturally human-understandable form. As a straightforward extension, we can chain multiple translations in the latent space and thus blend multiple concepts. Say that we have identified a set $M = \{C_i\}_{i=1}^m$ of concepts and found their corresponding CAVs, then we can add any linear combination of these concepts to the original point. 
\begin{equation}
    z' = z + \sum_{i=1}^m\varepsilon_i\hat{w}_{C_i}
\end{equation}
Thus the complicated task of modifying an existing design to exhibit a combination of multiple design concepts and styles has been reduced to a simple linear algebra operation.

\noindent \textbf{Concept querying} is a natural benefit of using the CAVs to interpret the latent space. Since the classifier essentially computes a similarity score between the CAV and the latent representation of the input, we can define the query with respect to a certain concept as 
\begin{equation}
    Q(\mathcal{D}_x,C) := \{q  : q = w_C \cdot e(x), x \in \mathcal{D}_x, \left|q\right| \gg 0\}.
\end{equation}
We can recover the instances in $\mathcal{D}_x$ that are most similar or dissimilar to the concept for strongly positive or negative $q$'s, respectively.

\section{Results}

In this section, we outline the results obtained for an in-house dataset consisting of 1165 car shapes represented by point clouds of dimension $\mathbb{R}^{6146 \times 3}$ with accompanying drag coefficients obtained from CFD simulations. A quarter of the shapes were reserved for validation, and the auto-encoder was trained on the rest. The auto-encoder consists of 4 fully connected layers for the encoder resulting in an embedding space with 8 dimensions; the decoder mirrors the encoder structure with the final output dimension equal to that of the original shapes. All activations in the auto-encoder are \texttt{LeakyReLU} \cite{Redmon2016YouOL} apart from the latent layer, which has a $tanh$ activation function to provide some hard bounds on the latent space, and no activation for the final layer of the decoder so that we can model the points in all of $\mathbb{R}^3$. For the regressor network $r_{\theta_r}$, we use a similar fully connected deep architecture to predict the drag coefficient for each car. Consequently we restrict the output to values in $\mathbb{R}_{>0}$ via a \texttt{ReLU} activation. Given the small size of the dataset, a small dropout was added to most layers to combat over-fitting. All models are trained for 1000 epochs using the ADAM optimizer \cite{Kingma2015AdamAM} with default settings. 

\subsection{Obtaining CAVs}

\begin{table}[t!]
  \centering
  \begin{tabular}{c c c}
    \toprule
    Concept & No. shapes & Example \\
    \midrule
    Sport & 9 & \raisebox{-.5\height}{\includegraphics[scale=0.1]{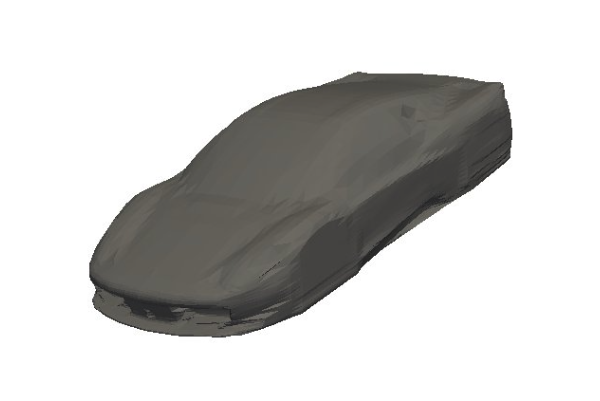}} \\
    Sedan & 9 & \raisebox{-.5\height}{\includegraphics[scale=0.1]{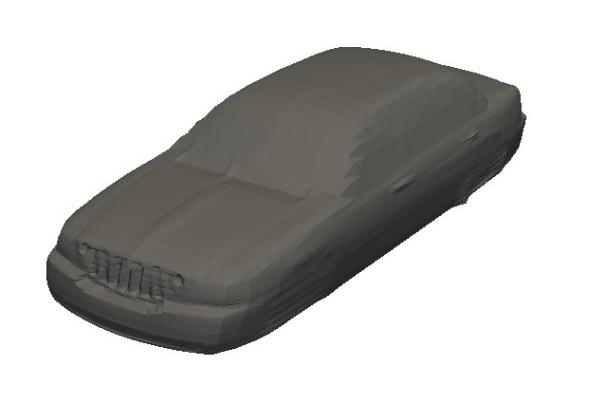}} \\
    Cuboids & 30 & \raisebox{-.5\height}{\includegraphics[scale=0.1]{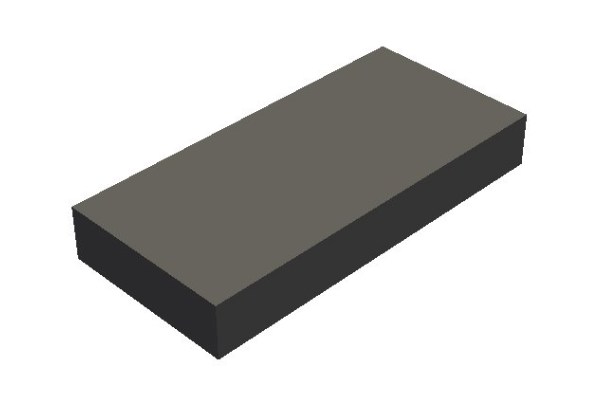}} \\
    Ellipsoids & 30 & \raisebox{-.5\height}{\includegraphics[scale=0.1]{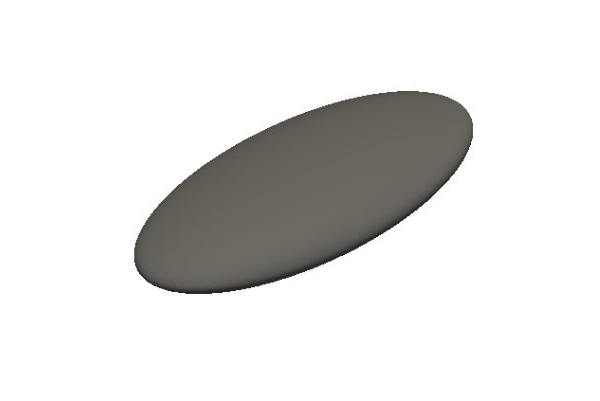}}\\
    \bottomrule
  \end{tabular}
  \caption{Collected and generated examples of concepts}
  \label{tab:selected}
\end{table}

\noindent \textbf{Collecting concepts}. To collect interesting concepts, we manually inspected the dataset and identified several styles of cars with a few examples each. In addition, to test the capability of the latent space to encode out of distribution shapes, we decided to represent the concept of boxiness and curviness via randomly generated cuboids and ellipsoids. To be close to the car shape distribution, we restrict the aspect ratio of the generated shapes to lie in a similar range to the cars. Specific numbers and examples are given in Table~\ref{tab:selected}. We chose the examples from the validation set to test the representation power of the latent space. Where non-concepts were necessary, we chose 50 random shapes from both the training and validation set.  

\begin{figure}[b!]
  \centering
  \includegraphics[width=0.85\linewidth,trim={0in 0in 0in 1in},clip]{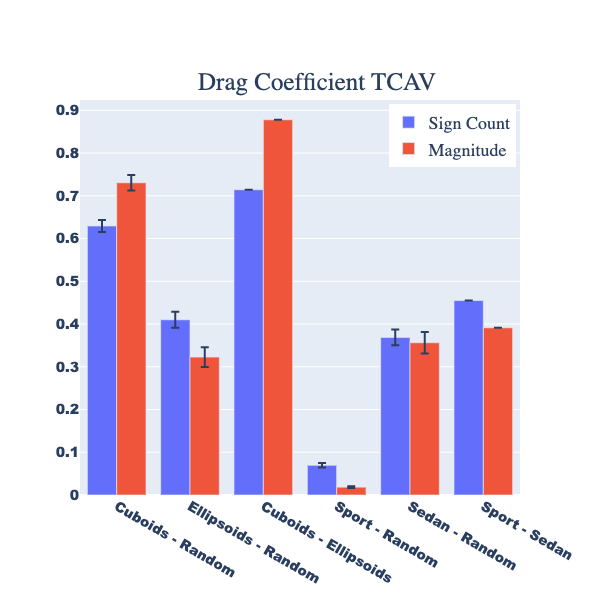}
  \caption{TCAV metric expressed in terms  of sign count and magnitude. Since the values for each concept in the CAV pair are complementary we only show the values for the first concept. Error bars represent one standard error.}
  \label{fig:drag_tcav}
\end{figure}

\begin{figure*}[t!]
    \centering
    \begin{subfigure}[b]{0.45\linewidth}
        \centering
        \includegraphics[width=0.95\linewidth, trim={0in 0in 0in 0.25in}, clip]{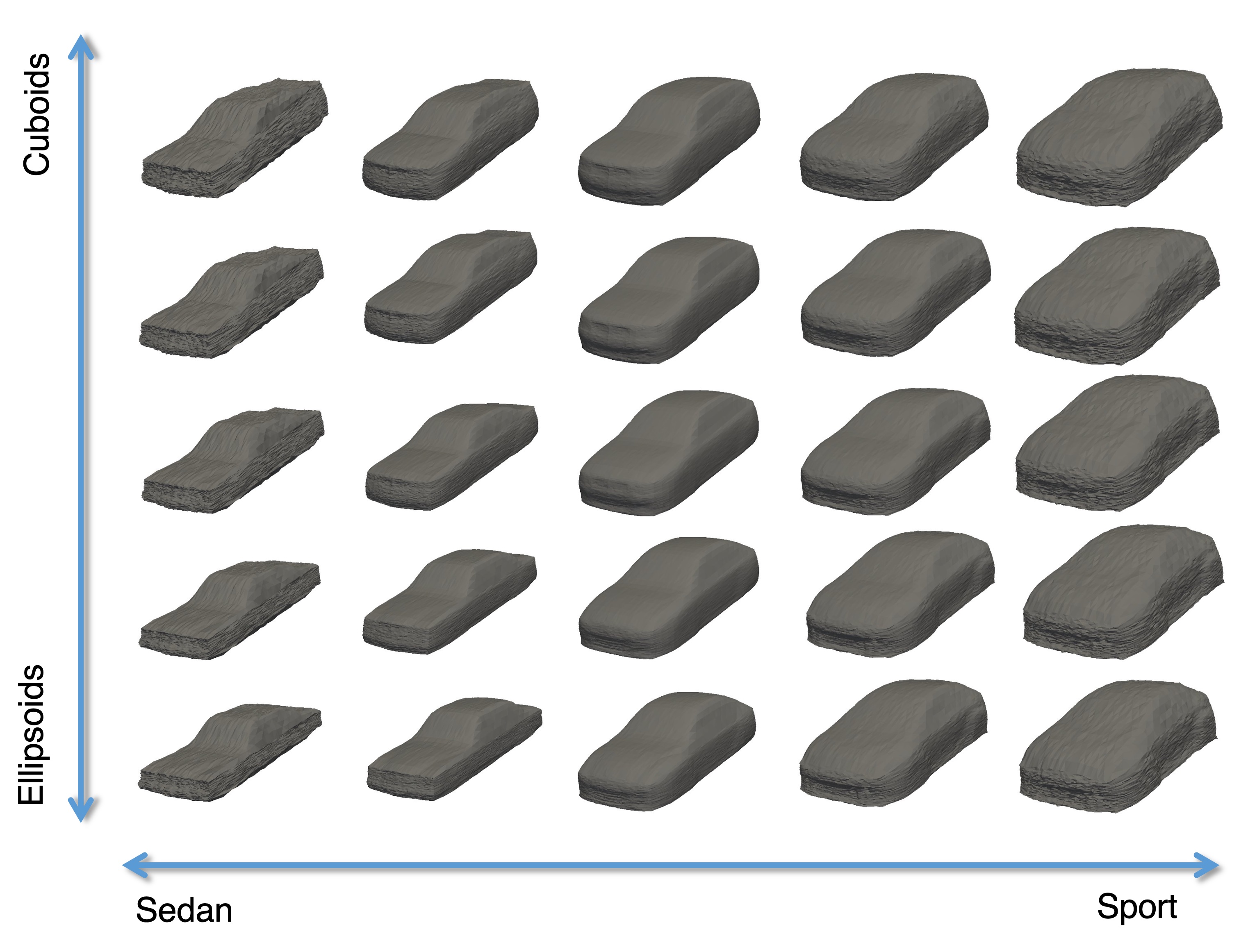}
        \label{fig:explore1}
    \end{subfigure}
    \hspace{0.2in}
    \begin{subfigure}[b]{0.45\linewidth}
        \centering
        \includegraphics[width=0.95\linewidth, trim={0in 0in 0in 0.25in}, clip]{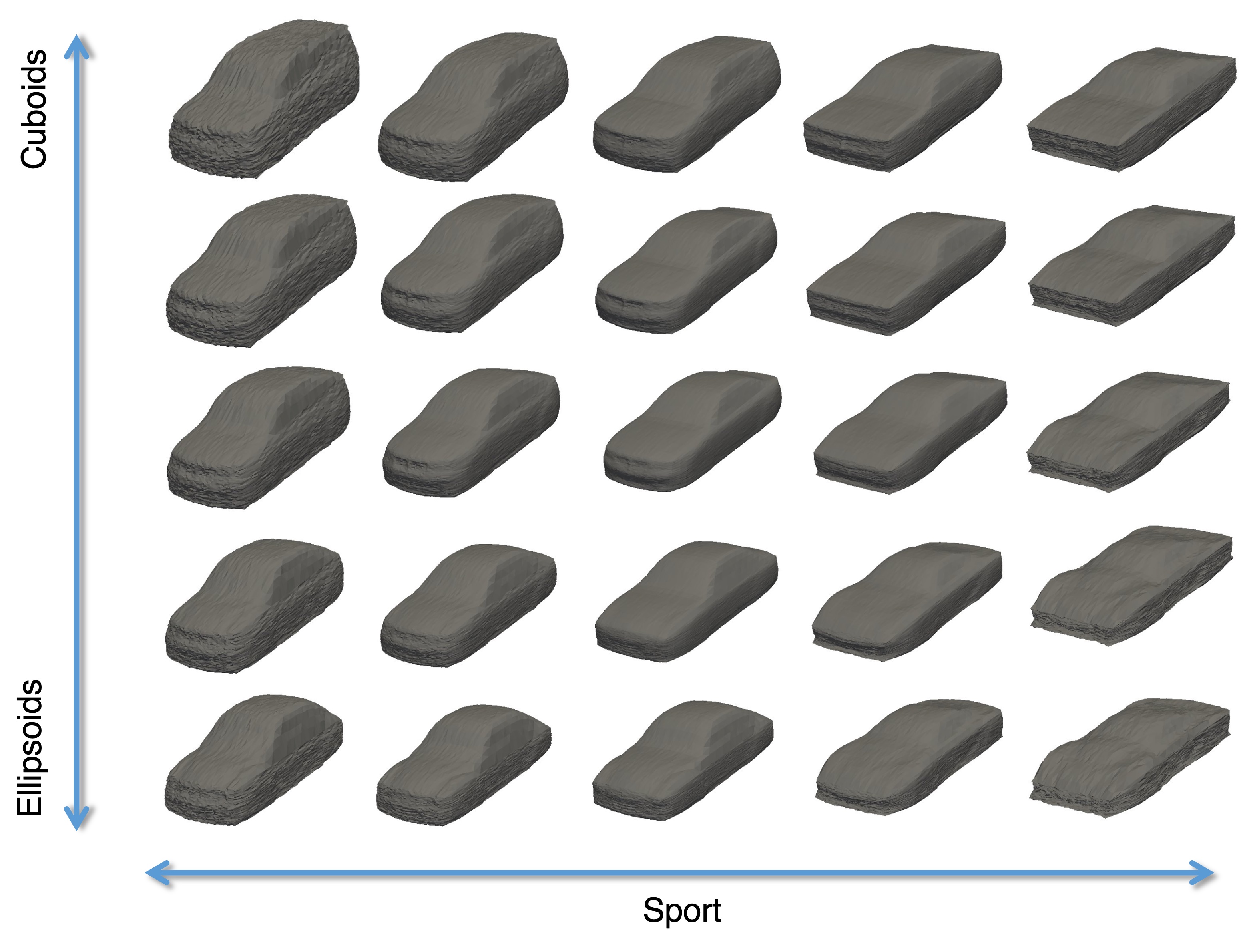}
        \label{fig:explore2}
    \end{subfigure}
    \caption{Concept blending for two pairs of CAVs. We blend the Ellipsoids - Cuboids CAV with the Sport - Sedan  CAV (left) and with the Sport - Random CAV (right). Note the different effect of moving towards the Sport concept in both cases.} 
    \label{fig:image_grids}
\end{figure*}

\noindent \textbf{Training CAVs}. Once the concept examples are selected, we train linear classifiers on the latent space representation of the examples via SGD and with an $L_2$ penalty. We explore CAVs obtained for each concept in Table~\ref{tab:selected} paired with a random subset of the car dataset. In addition, we train CAVs for the Cuboids-Ellipsoids and Sport-Sedan pairs since they are roughly opposite concepts. We can see the TCAV metrics of the drag coefficient in Figure~\ref{fig:drag_tcav}.

The TCAV metrics broadly follow our intuitions in that the less aero-dynamical concepts tend to increase the drag coefficients much more than the streamlined concepts. This is even more pronounced for the Cuboids-Ellipsoids CAV; however, the same can not be said for the Sport-Sedan CAV. This suggests that the Sedan concept might be more aerodynamic than first thought or that this is a poorly understood concept for the model. All the CAVs were verified to be statistically significant using a two-sided t-test.

\subsection{Concept Blending and Querying}

As discussed in Section~\ref{sec:exploring}, we can blend different concepts into an existing design by adding linear combinations of the CAVs to the latent space embeddings. Image grids for two pairs of interesting CAVs are presented in Figure~\ref{fig:image_grids}. First, we would like to note that both examples confirm the viability of our approach since the generated designs are not present in the original dataset and follow the intuition of blending designs well. In addition, it is interesting to observe how the behavior of the CAVs depends on the concepts used to generate them. Specifically moving towards the Sport concept for the Sport - Sedan CAV generates wider based and larger sporty designs in contrast to the Sport influenced shapes of the Sport - Random CAV. Both the synthetic concepts are also changing the design in sensible ways affecting both the shape of the front of the cars as well as the back to generate more compact or SUV-like types. Finally, the different concepts constrain each other in interesting ways depending on their relative strengths in the blend. For example, the last column of the right figure features muscle-type sports cars or racing-type sports cars depending on the degree of Cuboids or Ellipsoids added. 

\begin{figure}[t!]
  \centering
  \includegraphics[width=0.96\linewidth]{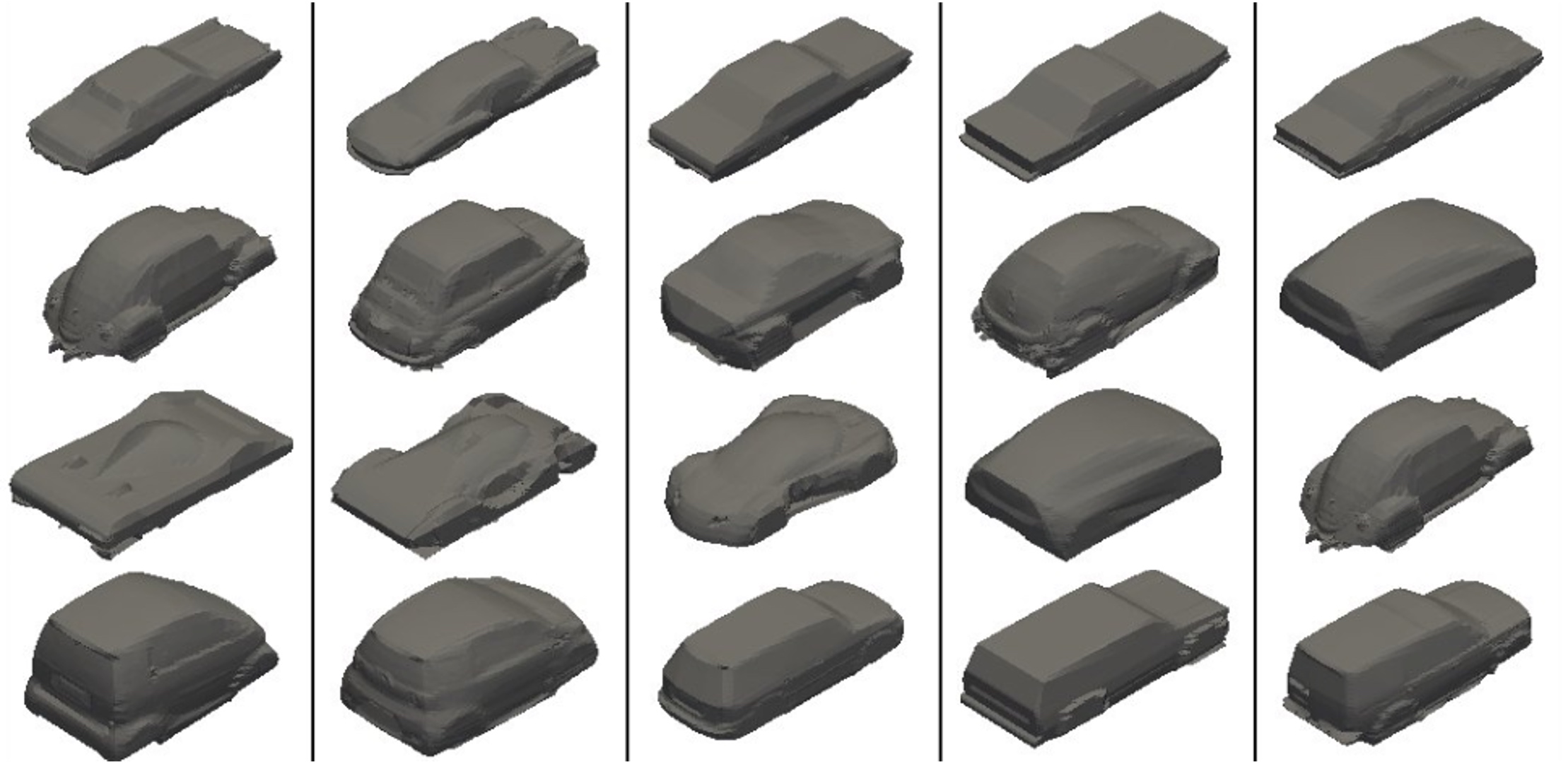}
  \caption{Top five queries from the dataset for a specific concept. From the top: Sedan, Ellipsoids, Sport and Cuboids.}
  \label{fig:querying}
\end{figure}

\begin{figure}[t!]
  \centering
  \includegraphics[width=.96\linewidth]{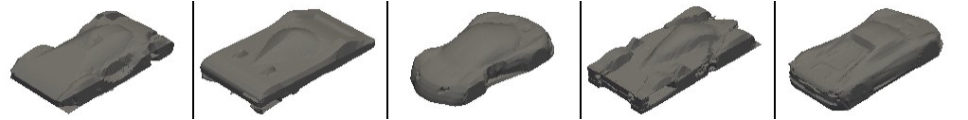}
  \caption{Query results for the re-computed Sport - Random CAV.}
  \label{fig:sports_q}
\end{figure}

\noindent \textbf{Querying}. We present the top five most similar designs for some concepts in Figure~\ref{fig:querying}. We note that querying the dataset is a good way to evaluate the quality of the learned CAV, as observed from the overlap between the results for the Sport and Ellipsoids concepts. Retraining the CAV for the Sport concept results in better query results in Figure~\ref{fig:sports_q}. This behavior emphasizes the stochastic nature of the CAV and that adding more examples and tuning the classifier might be worthwhile. 

\subsection{Parametric CAVs}

\begin{figure}[t!]
  \centering
  \raisebox{-.5\height}{\includegraphics[width=0.99\linewidth]{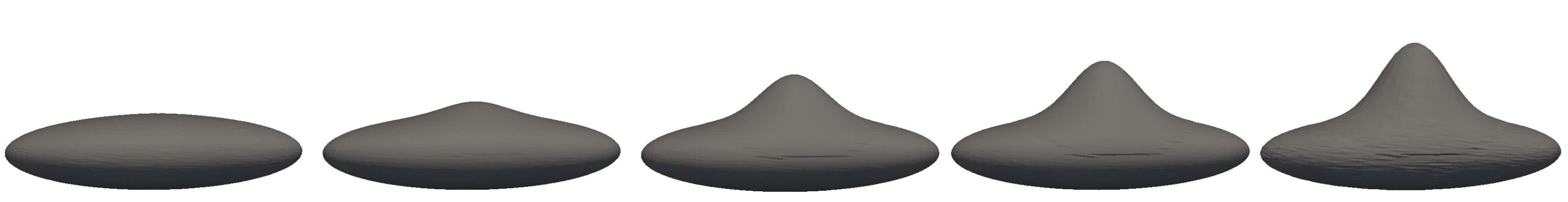}}
  \caption{Results of varying the HighBump CAV. Original shape is in the middle with negative $\varepsilon$ translations to the left and positive to the right.}
  \label{fig:fixed_swell}
\end{figure}

While the TCAV framework is great for exploring intangible concepts, it would also be useful for designers and engineers to control the parametric qualities of a design. To test whether this is possible, we generate a synthetic dataset of ellipsoids with constant proportions but with deformations of random height. We then select some examples with the highest deformations to obtain a HighBump - Random CAV. By varying the strength of the CAV, we can control the height of the deformation as seen in Figure~\ref{fig:fixed_swell}.

\begin{figure}[t!]
  \centering
  \includegraphics[width=0.99\linewidth]{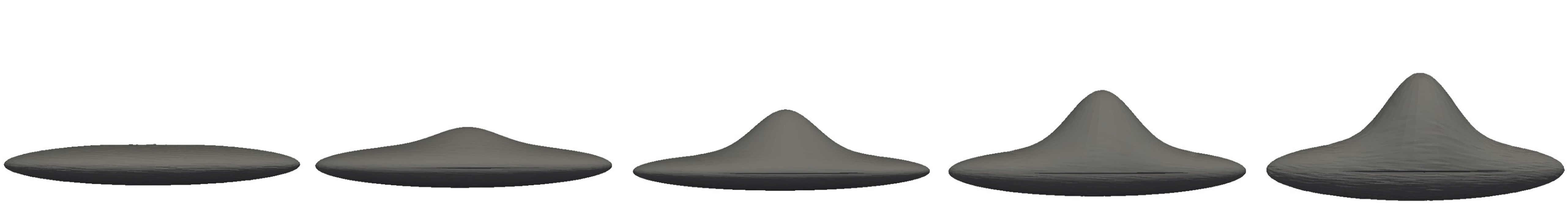}\\
  \vspace{0.3in}
  \includegraphics[width=0.99\linewidth]{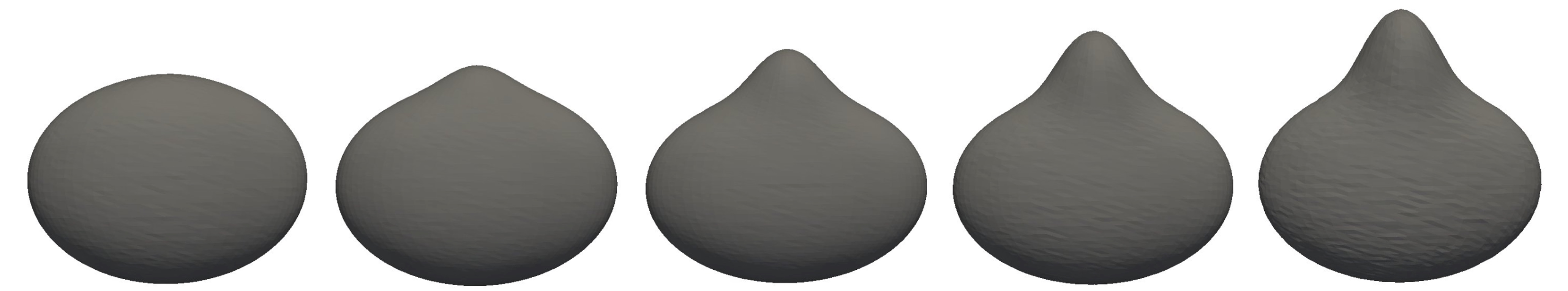}
  \caption{HighBump - LowBump CAV isolates the height parameter of the deformation for variable ellipsoid bodies.}
  \label{fig:random_swell}
\end{figure}

Even though the embedding dimension is set to 8, the data manifold is essentially one-dimensional, as evidenced by the mean off-diagonal correlation coefficient being 0.97. Thus it is not surprising that the CAV can control the height of the deformation so well. To determine if we could isolate the deformation from the other shape parameters, we trained another auto-encoder on a similar dataset but with random ellipsoid proportions. The new manifold is not one-dimensional with an absolute off-diagonal mean correlation of 0.57. However, because of the still significant correlation between the dimensions, we found it necessary to define the CAV using a converse LowBump concept instead of a non-concept. Using this CAV, we were able to vary the height of the deformation without affecting the rest of the ellipsoid shown in Figure~\ref{fig:random_swell}.

These results hint that the TCAV framework could be easily extended to regressive concepts. Indeed the HighBump - LowBump CAV is effectively a biased linear regression problem. 

\section{Discussion}

In this work, we demonstrated the applicability of the TCAV XAI framework to a dataset of 3D car shapes. We trained a simple auto-encoder to embed high dimensional point clouds into a low dimensional manifold. By gathering examples of different car types from the dataset together with abstract representations of style concepts, we were able to explore the latent space in a human interpretable way. Even with this simple architecture, the latent space proved to be rich enough to blend and represent concepts outside the data distribution. However, at the same time, the model was able to constrain the linear combinations of CAVs to produce interesting blends of various concepts without degenerating quickly into previously seen data points. 

We introduced the notion of a parametric CAV that allows for a reinterpretation of the latent space in terms of a known design parameter or one that might have become of interest later. Our approach was to cast the regression problem into a classification task to define the parametric CAV; however, we can use a linear regressor instead of a classifier. Using a linear regressor would, in principle, allow for the construction of single concept CAVs, and we leave it as a future direction of work. 

Finally, we would like to highlight that the high performance of the networks was not a prerequisite for the success of these experiments. Indeed, the low amount of data, simple architecture, and small latent dimension hampered the reconstruction capabilities of the auto-encoder. It is encouraging that concept blending worked well in this regime, and we believe it will also work well for high-performance, latent space-based generative models. TCAV may prove even more valuable for such models with a high number of latent dimensions since understanding and directly exploring these spaces is very challenging.  


\section{Acknowledgements} 
This work was done in the scope of the Innovate UK grant on "Explainable AI system to rationalise accelerated decision making on automotive component performance and manufacturability" (Project 10009522).

{\small
\bibliographystyle{ieee_fullname}
\bibliography{egbib}

\begin{thebibliography}{10}\itemsep=-1pt

\bibitem{Achlioptas2017RepresentationLA}
Panos Achlioptas, Olga Diamanti, Ioannis Mitliagkas, and Leonidas~J. Guibas.
\newblock Representation learning and adversarial generation of 3d point
  clouds.
\newblock {\em ArXiv}, abs/1707.02392, 2017.

\bibitem{adebayo2018sanity}
Julius Adebayo, Justin Gilmer, Michael Muelly, Ian Goodfellow, Moritz Hardt,
  and Been Kim.
\newblock Sanity checks for saliency maps.
\newblock {\em arXiv preprint arXiv:1810.03292}, 2018.

\bibitem{adebayo2020debugging}
Julius Adebayo, Michael Muelly, Ilaria Liccardi, and Been Kim.
\newblock Debugging tests for model explanations.
\newblock {\em arXiv preprint arXiv:2011.05429}, 2020.

\bibitem{atzmon2020sal}
Matan Atzmon and Yaron Lipman.
\newblock {SAL}: Sign agnostic learning of shapes from raw data.
\newblock In {\em Proceedings of the IEEE/CVF Conference on Computer Vision and
  Pattern Recognition}, pages 2565--2574, 2020.

\bibitem{atzmon2020sald}
Matan Atzmon and Yaron Lipman.
\newblock Sald: Sign agnostic learning with derivatives.
\newblock {\em arXiv preprint arXiv:2006.05400}, 2020.

\bibitem{binder2016layer}
Alexander Binder, Gr{\'e}goire Montavon, Sebastian Lapuschkin, Klaus-Robert
  M{\"u}ller, and Wojciech Samek.
\newblock Layer-wise relevance propagation for neural networks with local
  renormalization layers.
\newblock In {\em International Conference on Artificial Neural Networks},
  pages 63--71. Springer, 2016.

\bibitem{Brock2016GenerativeAD}
Andrew Brock, Theodore Lim, James~M. Ritchie, and Nick Weston.
\newblock Generative and discriminative voxel modeling with convolutional
  neural networks.
\newblock {\em ArXiv}, abs/1608.04236, 2016.

\bibitem{chen2020bsp}
Zhiqin Chen, Andrea Tagliasacchi, and Hao Zhang.
\newblock Bsp-net: Generating compact meshes via binary space partitioning.
\newblock In {\em Proceedings of the IEEE/CVF Conference on Computer Vision and
  Pattern Recognition}, pages 45--54, 2020.

\bibitem{Chen2019LearningIF}
Zhiqin Chen and Hao Zhang.
\newblock Learning implicit fields for generative shape modeling.
\newblock {\em 2019 IEEE/CVF Conference on Computer Vision and Pattern
  Recognition (CVPR)}, pages 5932--5941, 2019.

\bibitem{deng2020cvxnet}
Boyang Deng, Kyle Genova, Soroosh Yazdani, Sofien Bouaziz, Geoffrey Hinton, and
  Andrea Tagliasacchi.
\newblock Cvxnet: Learnable convex decomposition.
\newblock In {\em Proceedings of the IEEE/CVF Conference on Computer Vision and
  Pattern Recognition}, pages 31--44, 2020.

\bibitem{Donahue2017AdversarialFL}
Jeff Donahue, Philipp Kr{\"a}henb{\"u}hl, and Trevor Darrell.
\newblock Adversarial feature learning.
\newblock {\em ArXiv}, abs/1605.09782, 2017.

\bibitem{ghadai2018learning}
Sambit Ghadai, Aditya Balu, Soumik Sarkar, and Adarsh Krishnamurthy.
\newblock Learning localized features in {3D CAD} models for manufacturability
  analysis of drilled holes.
\newblock {\em Computer Aided Geometric Design}, 62:263--275, 2018.

\bibitem{goyal2019explaining}
Yash Goyal, Amir Feder, Uri Shalit, and Been Kim.
\newblock Explaining classifiers with causal concept effect (cace).
\newblock {\em arXiv preprint arXiv:1907.07165}, 2019.

\bibitem{gropp2020implicit}
Amos Gropp, Lior Yariv, Niv Haim, Matan Atzmon, and Yaron Lipman.
\newblock Implicit geometric regularization for learning shapes.
\newblock {\em arXiv preprint arXiv:2002.10099}, 2020.

\bibitem{Guan2020GeneralizedAF}
Yanran Guan, Tansin Jahan, and Oliver~Matias van Kaick.
\newblock Generalized autoencoder for volumetric shape generation.
\newblock {\em 2020 IEEE/CVF Conference on Computer Vision and Pattern
  Recognition Workshops (CVPRW)}, pages 1082--1088, 2020.

\bibitem{huang2019claim}
Shikun Huang, Binbin Zhang, Wen Shen, and Zhihua Wei.
\newblock A claim approach to understanding the pointnet.
\newblock In {\em Proceedings of the 2019 2nd International Conference on
  Algorithms, Computing and Artificial Intelligence}, pages 97--103, 2019.

\bibitem{kanezaki2016rotationnet}
Asako Kanezaki, Yasuyuki Matsushita, and Yoshifumi Nishida.
\newblock {RotationNet}: {J}oint object categorization and pose estimation
  using multiviews from unsupervised viewpoints.
\newblock In {\em Proceedings of the IEEE Conference on Computer Vision and
  Pattern Recognition}, pages 5010--5019, 2018.

\bibitem{kim2018interpretability}
Been Kim, Martin Wattenberg, Justin Gilmer, Carrie Cai, James Wexler, Fernanda
  Viegas, et~al.
\newblock Interpretability beyond feature attribution: Quantitative testing
  with concept activation vectors (tcav).
\newblock In {\em International conference on machine learning}, pages
  2668--2677. PMLR, 2018.

\bibitem{Kingma2015AdamAM}
Diederik~P. Kingma and Jimmy Ba.
\newblock Adam: A method for stochastic optimization.
\newblock {\em CoRR}, abs/1412.6980, 2015.

\bibitem{Kudo2019VirtualTS}
Akira Kudo, Y. Kitamura, Yuanzhong Li, Satoshi Iizuka, and Edgar Simo-Serra.
\newblock Virtual thin slice: 3d conditional gan-based super-resolution for ct
  slice interval.
\newblock In {\em MLMIR@MICCAI}, 2019.

\bibitem{lecun2015deep}
Yann LeCun, Yoshua Bengio, and Geoffrey Hinton.
\newblock Deep learning.
\newblock {\em nature}, 521(7553):436--444, 2015.

\bibitem{li2018sonet}
Jiaxin Li, Ben~M Chen, and Gim~Hee Lee.
\newblock {So-net}: {S}elf-organizing network for point cloud analysis.
\newblock In {\em Proceedings of the IEEE Conference on Computer Vision and
  Pattern Recognition}, pages 9397--9406, 2018.

\bibitem{Lin20163DKD}
Xinyu Lin, Ce Zhu, Q. Zhang, and Y. Liu.
\newblock 3d keypoint detection based on deep neural network with sparse
  autoencoder.
\newblock {\em ArXiv}, abs/1605.00129, 2016.

\bibitem{Mildenhall2020NeRFRS}
Ben Mildenhall, Pratul~P. Srinivasan, Matthew Tancik, Jonathan~T. Barron, Ravi
  Ramamoorthi, and Ren Ng.
\newblock Nerf: Representing scenes as neural radiance fields for view
  synthesis.
\newblock In {\em ECCV}, 2020.

\bibitem{Mller2022InstantNG}
Thomas M{\"u}ller, Alex Evans, Christoph Schied, and Alexander Keller.
\newblock Instant neural graphics primitives with a multiresolution hash
  encoding.
\newblock {\em ArXiv}, abs/2201.05989, 2022.

\bibitem{Nash2017TheSV}
Charlie Nash and Christopher K.~I. Williams.
\newblock The shape variational autoencoder: A deep generative model of
  part‐segmented 3d objects.
\newblock {\em Computer Graphics Forum}, 36, 2017.

\bibitem{Park2019DeepSDFLC}
Jeong~Joon Park, Peter~R. Florence, Julian Straub, Richard~A. Newcombe, and S.
  Lovegrove.
\newblock Deepsdf: Learning continuous signed distance functions for shape
  representation.
\newblock {\em 2019 IEEE/CVF Conference on Computer Vision and Pattern
  Recognition (CVPR)}, pages 165--174, 2019.

\bibitem{paschalidou2021neural}
Despoina Paschalidou, Angelos Katharopoulos, Andreas Geiger, and Sanja Fidler.
\newblock Neural parts: Learning expressive 3d shape abstractions with
  invertible neural networks.
\newblock In {\em Proceedings of the IEEE/CVF Conference on Computer Vision and
  Pattern Recognition}, pages 3204--3215, 2021.

\bibitem{qi2016volumetric}
Charles~R Qi, Hao Su, Matthias Nie{\ss}ner, Angela Dai, Mengyuan Yan, and
  Leonidas~J Guibas.
\newblock Volumetric and multi-view {CNNs} for object classification on {3D}
  data.
\newblock In {\em Proceedings of the IEEE Conference on Computer Vision and
  Pattern Recognition}, pages 5648--5656, 2016.

\bibitem{Redmon2016YouOL}
Joseph Redmon, Santosh~Kumar Divvala, Ross~B. Girshick, and Ali Farhadi.
\newblock You only look once: Unified, real-time object detection.
\newblock {\em 2016 IEEE Conference on Computer Vision and Pattern Recognition
  (CVPR)}, pages 779--788, 2016.

\bibitem{selvaraju2017grad}
Ramprasaath~R Selvaraju, Michael Cogswell, Abhishek Das, Ramakrishna Vedantam,
  Devi Parikh, and Dhruv Batra.
\newblock Grad-cam: Visual explanations from deep networks via gradient-based
  localization.
\newblock In {\em Proceedings of the IEEE international conference on computer
  vision}, pages 618--626, 2017.

\bibitem{SFIKAS2018208}
Konstantinos Sfikas, Ioannis Pratikakis, and Theoharis Theoharis.
\newblock Ensemble of {PANORAMA}-based convolutional neural networks for {3D}
  model classification and retrieval.
\newblock {\em Computers \& Graphics}, 2017.

\bibitem{simonyan2013deep}
Karen Simonyan, Andrea Vedaldi, and Andrew Zisserman.
\newblock Deep inside convolutional networks: Visualising image classification
  models and saliency maps.
\newblock {\em arXiv preprint arXiv:1312.6034}, 2013.

\bibitem{Sitzmann2020ImplicitNR}
Vincent Sitzmann, Julien N.~P. Martel, Alexander~W. Bergman, David~B. Lindell,
  and Gordon Wetzstein.
\newblock Implicit neural representations with periodic activation functions.
\newblock {\em ArXiv}, abs/2006.09661, 2020.

\bibitem{song2016deep}
Shuran Song and Jianxiong Xiao.
\newblock Deep sliding shapes for amodal {3D} object detection in {RGB-D}
  images.
\newblock In {\em Proceedings of the IEEE Conference on Computer Vision and
  Pattern Recognition}, pages 808--816, 2016.

\bibitem{su2021learning}
Feng-Guang Su, Ci-Siang Lin, and Yu-Chiang~Frank Wang.
\newblock Learning interpretable representation for 3d point clouds.
\newblock In {\em 2020 25th International Conference on Pattern Recognition
  (ICPR)}, pages 7470--7477. IEEE, 2021.

\bibitem{su15mvcnn}
Hang Su, Subhransu Maji, Evangelos Kalogerakis, and Erik Learned-Miller.
\newblock Multi-view convolutional neural networks for {3D} shape recognition.
\newblock In {\em Proceedings of the IEEE International Conference on Computer
  Vision}, pages 945--953, 2015.

\bibitem{tan2022surrogate}
Hanxiao Tan and Helena Kotthaus.
\newblock Surrogate model-based explainability methods for point cloud nns.
\newblock In {\em Proceedings of the IEEE/CVF Winter Conference on Applications
  of Computer Vision}, pages 2239--2248, 2022.

\bibitem{tomsett2020sanity}
Richard Tomsett, Dan Harborne, Supriyo Chakraborty, Prudhvi Gurram, and Alun
  Preece.
\newblock Sanity checks for saliency metrics.
\newblock In {\em Proceedings of the AAAI conference on artificial
  intelligence}, volume 34-04, pages 6021--6029, 2020.

\bibitem{Wu2016LearningAP}
Jiajun Wu, Chengkai Zhang, Tianfan Xue, Bill Freeman, and Joshua~B. Tenenbaum.
\newblock Learning a probabilistic latent space of object shapes via 3d
  generative-adversarial modeling.
\newblock In {\em NIPS}, 2016.

\bibitem{wu20153d}
Zhirong Wu, Shuran Song, Aditya Khosla, Fisher Yu, Linguang Zhang, Xiaoou Tang,
  and Jianxiong Xiao.
\newblock {3D ShapeNets}: {A} deep representation for volumetric shapes.
\newblock In {\em Proceedings of the IEEE Conference on Computer Vision and
  Pattern Recognition}, pages 1912--1920, 2015.

\bibitem{Yin2018P2PNET}
K. Yin, Hui Huang, Daniel Cohen-Or, and Hao Zhang.
\newblock P2p-net.
\newblock {\em ACM Transactions on Graphics (TOG)}, 37:1 -- 13, 2018.

\bibitem{yoo2021explainable}
Soyoung Yoo and Namwoo Kang.
\newblock Explainable artificial intelligence for manufacturing cost estimation
  and machining feature visualization.
\newblock {\em Expert Systems with Applications}, 183:115430, 2021.

\bibitem{zhang2019explaining}
Binbin Zhang, Shikun Huang, Wen Shen, and Zhihua Wei.
\newblock Explaining the pointnet: What has been learned inside the pointnet?
\newblock In {\em CVPR Workshops}, pages 71--74, 2019.

\bibitem{zhou2016learning}
Bolei Zhou, Aditya Khosla, Agata Lapedriza, Aude Oliva, and Antonio Torralba.
\newblock Learning deep features for discriminative localization.
\newblock In {\em Proceedings of the IEEE conference on computer vision and
  pattern recognition}, pages 2921--2929, 2016.

\end{thebibliography}
}

\end{document}